\documentclass{article}

    \usepackage[preprint]{neurips_2022}

\usepackage[utf8]{inputenc} 
\usepackage[T1]{fontenc}    
\usepackage{hyperref}       
\usepackage{url}            
\usepackage{booktabs}       
\usepackage{amsfonts}       
\usepackage{nicefrac}       
\usepackage{microtype}      
\usepackage{xcolor}         
\usepackage{epsfig,epstopdf,algorithmic,algorithm,textcomp,amsthm,multirow,moresize,amsfonts}
\usepackage{multicol}
\newtheorem{theorem}{Theorem}

\newtheorem{corollary}{Corollary}

\newtheorem{assumption}{Assumption}
\newtheorem{proposition}{Proposition}

\newtheorem{definition}{Definition}
\newtheorem{remark}{Remark}
\usepackage{multirow}
\usepackage{amsmath,bm}
\usepackage{breqn}
\DeclareMathOperator*{\argmin}{argmin}
\newcommand{\ind}{\perp\!\!\!\!\perp}

\title{Pairwise Learning via Stagewise Training in Proximal Setting}

\author{%
  Hilal AlQuabeh \quad Aliakbar Abdurahimov \\
  Department of Machine Learning\\
  Mohamed bin Zayed University of Artificial Intelligence\\
  UAE, Abu Dhabi \\
  \texttt{hilal.alquabeh,aliakbar.abdurahimov@mbzuai.ac.ae}
\\
}

\begin{document}

\maketitle

\begin{abstract}
The pairwise objective paradigms are an important and essential aspect of machine learning.
Examples of machine learning approaches that use pairwise objective functions include differential network in face recognition, metric learning, bipartite learning, multiple kernel learning, and maximizing of area under the curve (AUC). Compared to pointwise learning, pairwise learning's sample size grows quadratically with the number of samples and thus its complexity.
Researchers mostly address this challenge by utilizing an online learning system. Recent research has, however, offered adaptive sample size training for smooth loss functions as a better strategy in terms of convergence and complexity, but without a comprehensive theoretical study. 
In a distinct line of research, importance sampling has sparked a considerable amount of interest in finite pointwise-sum minimization. This is because of the stochastic gradient variance, which causes the convergence to be slowed considerably. 
In this paper, we combine adaptive sample size and importance sampling techniques for pairwise learning, with convergence guarantees for nonsmooth convex pairwise loss functions. In particular, the model is trained stochastically using an expanded training set for a predefined number of iterations derived from the stability bounds.
In addition, we demonstrate that sampling opposite instances at each iteration reduces the variance of the gradient, hence accelerating convergence. 
Experiments on a broad variety of datasets in AUC maximization confirm the theoretical results. 
\end{abstract}

\section{Introduction}
In machine learning, an unknown optimal classifier is sought to minimize the overall true or excess error associated with classifying all underlying distributions.
Since the distribution is not completely known, sampling statistics are used to depict it, for instance, empirical loss (finite sum of pointwise losses) is proved to be an unbiased estimate of the true risk. However, in many situations, such as imbalanced data classification, where labels are not evenly distributed across the training set,  continuing to approach the learning process in the classical pointwise empirical risk minimization may result in enormous computational effort.
In the latter scenario and others e.g. differential network in face recognition \cite{kang2018pairwise, song2019occlusion}, metric learning \cite{kulis2012metric}, bipartite learning \cite{zhang2016pairwise}, multiple kernel learning \cite{zhuang2011unsupervised}, and area under the curve (AUC) maximization \cite{zhao2011online},
 the pairwise objective paradigms is one alternative with superior statistical properties.
\par
In contrast to pointwise learning, the sample size for pairwise learning increases quadratically as the number of samples increases.
Researchers address this challenge by adopting an online learning scheme. For example Boissier et al. \cite {boissier2016fast} introduced an improved online algorithm for
 general pairwise learning with complexity of $O(Td^2)$ \footnote{The complexity is measured by gradient components evaluations where T is iteration number and d is the dimension of the data}. In AUC maximization, Zhao et al. employs reservoir sampling with size $s$, to sidestep the pairwise quadratic complexity in an online framework to enhance the complexity to $O(sTd)$ \cite{zhao2011online}. Recently, Ying et al. \cite{ying2016stochastic}, have obtained an equivalent formulation of the AUC convex square linear loss using the saddle point problem and have optimized it in a stochastic manner to enhance the complexity to $O(Td)$. Based on saddle point formulation of AUC objective, 
Natole et al.\cite{pmlr-v80-natole18a} enhanced the algorithm such that extra variables introduced in \cite{ying2016stochastic} are found in closed form solutions, which enhances the convergence of the original work from $\mathcal{O}(\frac{1}{\sqrt{T}})$ to $\mathcal{O}(\frac{1}{T})$ under strong convexity assumption of the objective function but kept same complexity.  
\par 
Moreover, many offline algorithms were proposed in the pairwise framework, for instance, Yang et al. recently investigated the offline stochastic gradient descent for pairwise learning in which at every iteration one pair is randomly selected from uniform distribution over $[n]$ i.e. the number of available examples. \cite{yang2021simple}. Gu at al. \cite{gu2019scalable} implemented doubly stochastic gradient descent to update a linear model incrementally in an stagewise offline manner, where again prior uniform distribution is assumed on the training set every stage.
As with point-wise stochastic gradient descent, however, randomly picking one pair and modeling the gradient solely on one pair generates inevitable variance that slows convergence.
\par
Although importance sampling is a fundamental strategy for dealing with variance created in stochastic learning systems, none of the available research has studied the importance  sampling in pairwise paradigm to the best of our knowledge.  
On the other hand, we make advantage of the successful analysis presented in \cite{daneshmand2016starting} with regard to the computational complexity in connection to the generalization and optimization errors. In particular, we tackle the quadratic complexity of finite sum minimization by employing an established yet effective adaptive sample size technique in which the problem is splitted to multi sub problems with expanding sample size. Although the method has been studied in the context of pointwise \cite{mokhtari2016adaptive} smooth-pariwise recently \cite{gu2019scalable}, but none of the researches tackle the non-smooth pairwise optimization. Moreover, we analyze the sub problems iteration number from different approach, namely the uniform stability and the optimization error as discussed in section \ref{Methodology}.
 
 \begin{table}[!h]
 \centering
 \label{table:comp}
 \caption{Recent pairwise learning algorithms( T is the iteration number, d: the dimension and s is a buffer size).}
\begin{tabular}{lllllll}
\hline
\textbf{Algorithm} & \textbf{Reference} & \textbf{Problem} & \textbf{Loss} & \textbf{Sampling} & \textbf{Complexity}  \\ \hline
SPAM-NET           &      \cite{pmlr-v80-natole18a}              & AUC              & NS-CVX        & Online            & $O(Td)$                             \\
SOLAM              &           \cite{ying2016stochastic}         & AUC              & S-CVX         & Online            & $O(Td)$                          \\
OAM                &        \cite{zhao2011online}            & AUC              & S-CVX         & Online            & $O(sTd)$              \\
OPAUC              &        \cite{gao2013one}            & AUC              & S-CVX         & Online            &           $O(n)$            \\
AdaDSGD             &      \cite{gu2019scalable}              & General          & S-CVX         & Uniform           &           $O(n)$             \\ \hline
adaPSGD            &        Ours            & Genereal         & NS-CVX        & Opposite          & $O(Td)$                             \\ \hline
\end{tabular}
\end{table}

\subsection{Related work}
The adaptive sample size approach is investigated in literature in different frameworks. The fundamental stochastic gradient descent (SGD) algorithm is one of the early techniques taking advantages of the identically distributed  and independently drawn (i.i.d.)  finite sum structure of the ERM. Followed by that the pioneer work of SAGA \cite{defazio2014saga}, SVRG\cite{reddi2016stochastic}, SARAH\cite{nguyen2017sarah} and ADAM\cite{kingma2014adam}. The primary motivation for all of the above techniques is to decrease the variance caused by randomly selecting a fixed-size subset every iteration. Recently Daneshmand et al. implemented an adaptive size scheme of training samples every iteration, and proved a better computational complexity in new algorithm (dynSAGA) to have statistical accurate solution in $log(n)n$ instead of $2n$ iterations\cite{daneshmand2016starting}. Moreover, Mokhtari et al. \cite{mokhtari2017first} applied  sample size methodology to first-order stochastic and deterministic algorithms, notably SVRG and accelerated gradient descent, and later to newton method \cite{mokhtari2016adaptive} where better computational complexity is achieved. However in all mentioned researches, only univariate ERM or regularized ERM.
Different authors are addressing pairwise learning in which full gradient needs a visit to all possible pairs i.e. $O(n^2)$, where $n$ refers to the number of training samples and thus deterministic approach serves as naive upper bound of complexity in $O(Tn^2)$ where $T$ is iterations number. Zhao et al. presented an online approach for AUC maximization \cite{zhao2011online} where buffer sampling with size $S$ is implemented to reduce computational complexity to $O(Sd)$ with squared pairwise loss and Frobenius norm, with d being the dimension of data. Gau et al. \cite{gao2013one} presented covariance matrix to update the gradient in $O(Td^2)$, however this approach is very infeassible in high-dimensional data. Ying et al. \cite{ying2016stochastic} converted the pairwise loss function with $l_2$ regularization to saddle point point problem, and update the primal and dual variables stochastically with $O(Td)$ complexity. The saddle point problem of AUC is also investigated in \cite{pmlr-v80-natole18a} with non-smooth regularization namely, elastic net, with same computational complexity but better convergence rate under strong convexity setting. Although the work of Natole et al. \cite{pmlr-v80-natole18a} handles the non-smooth regularization, it only focuess on the AUC maximization but we condiser the pairwise learning in general. Recently, Gu et al. \cite{gu2019scalable} performed the adaptive sample size scheme on pairwise loss function, where experiment on AUC proved a superior results in terms of convergence $O(1/T)$ and computational complexity $O(Td)$ ,  as shown in table \ref{table:comp}.  This research is most relevant to the last research, but with non-smooth regularized ERM and room of imporovment in the sub problem iteration number.
\subsection{Contributions}
We focus on pairwise learning in general, and AUC in particular, in this study, and make the following contributions: 
\begin{itemize}
    \item  Design new algorithm for non-smooth pairwise learning with adaptive sample size, and non-uniform sampling technique.
    \item Analyze the sub problems iteration number from uniform stability and convergence analysis.
\end{itemize}
The remainder of the study is structured as follows: section \ref{sec:prb}
defines the problem of research, section 
\ref{Methodology} presents the methods for solving pairwise learning, section \ref{sec:exp} illustrates the experiments, and section \ref{sec:con} concludes with closing remarks. 

\section{Problem Definition}
\label{sec:prb}
In pairwise learning the ultimate goal is to learn a hypothesis over some distribution $\mathcal{D}:=(\mathcal{X},\mathcal{Y},\mathbb{P})$ where $\mathcal{X}$ is input space, $\mathcal{Y}$ is output space, and $\mathbb{P}$ is a probability distribution. In cotrast to pointwise learning, in pairwise learning samples are available as pairs of samples and thus a pairwise loss function is implemented. In particular assumes a space $\mathcal{Z} =\mathcal{X}\times \mathcal{Y} $ and a pairwise loss function of hypothesis $w\in \mathbb{R}^d$ as $\ell : \mathbb{R}^d \times \mathcal{Z}\times \mathcal{Z} \rightarrow \mathbb{R}_+$. Some examples of pairwise loss functions are squared loss function $\ell(w,(x,y),(x',y')=(1-f(x)-f(x'))^2$, and hinge loss $\ell(w,(x,y),(x',y') = \max(0,1-f(x)+f(x'))$ where $y = 1$, $y'=0$ and $f(\cdot)$ is the hypothesis mapping. Moreover, the primary goal is to minimize an expected loss (risk) which is defined as $L(w) :=\mathbb{E}_{z\sim \mathcal{D}} \mathbb{E}_{z'\sim \mathcal{D}} [\ell(w,z,z')]$ or the regularized expected loss as illustrated in \ref{eq:risk}:
	\begin{equation}
	\label{eq:risk}
	R(w):=\underbrace{\mathbb{E}_{z\sim \mathcal{D}} \mathbb{E}_{z'\sim \mathcal{D}} [\ell(w,z,z')]}_{L(w)} + \lambda \Omega(w)
	\end{equation} 
	where $\Omega:\mathbb{R}^d \rightarrow \mathbb{R}_+$ is possibly non-smooth regularizer and $\lambda$ is the regularization parameter. Formulation in \ref{eq:risk} is common in machine learning e.g. AUC maximization \cite{ying2016stochastic}.  
	 Since the expected loss is built on the unknown distributions $\mathcal{D}$, the empirical risk given a samples of data $\mathcal{S} \sim \mathcal{D} ^m$ has been demonstrated to be a good estimator of the expected loss as illustrated in \ref{eq:ERM}.

\begin{equation}
\label{eq:ERM}
R_n(w) := \underbrace{\frac{1}{n(n-1)} \sum_{i\neq j}\ell(w,z_i,z_j)}_{L_n(w)}  + \lambda \Omega(w)
\end{equation}

\subsection{Assumptions and Notations}
\label{assumptions}
Given $\mathcal{X}\subset \mathbb{R}^d$,$\mathcal{Y} = \{0,1\}$ being some input and output space, and unknown distributions $\mathcal{D}_+ = \mathbb{P}_{(x,y)\sim \mathcal{D}_+}(x,y=1)$ and $\mathcal{D}_- = \mathbb{P}_{(x,y)\sim \mathcal{D}_-}(x,y=0)$ on $\mathcal{X}\times \mathcal{Y}$, we first consider a training subset $$S:= \{z_1=(x_1,y_1),\dots,z_m=(x_n,y_n) \}$$ where $|S|:=n$ and $S$ is i.i.d of $x_i \in \mathcal{X}$ and $y_i\in \mathcal{Y}$ i.e. $\{z_i\}_{i=1}^n\sim \mathcal{D}^n$,  where $$\mathcal{D}(z) = \alpha \mathcal{D}_+(z) + (1-\alpha) \mathcal{D}_-(z) \quad \forall z,\alpha\footnote{if $\alpha =0.5$ then we have equal portion of positive and negatives examples in $\mathcal{S}$}\in \mathcal{X}\times \mathcal{Y}\times [0,1]$$ 
We also define a modified training subset of $\mathcal{S}$  as follow,
\begin{eqnarray}
\mathcal{S}^i:= \{z_1,\dots,z^i,\dots,z_n \}
\label{eq:modset}
\end{eqnarray}
where $z^i \sim \mathcal{D}$ and independent from other examples i.e. $z^i \ind \mathcal{S}$.

We list all assumptions as follow assuming linear hypothesis e.g. $f(w,x):=w^Tx$
\begin{assumption}[Lipschits Continuity] 
	\label{ass:Lipschits} For any $ z,z' \in \mathcal{X}$ the loss function $\ell(\cdot;z,z')$ is G-Lipschits continuous function over the feasible set $\mathbb{R}^d$ i.e. $$ |\ell(w;z,z') - \ell(w';z,z')|\leq G \|w-w' \| $$.
	\label{ass:lips}
\end{assumption}
\begin{assumption}[Convexity] 
	\label{ass:convex} For any $ z,z' \in \mathcal{X}$ the loss function $\ell(\cdot;z,z')$ is $\mu$-strongly convex function over the feasible set $\mathbb{R}^d$ i.e. $$ \ell(w;z,z') \geq \ell(w';z,z') + \nabla \ell(w';z,z')^(w-w')+\frac{\mu}{2}\|w'-w\|^2$$ moreover, the possibly non-smooth function $\Omega :\mathbb{R}^d \rightarrow \mathbb{R}_+$ is convex. Thus we have that $L(w;z,z') :=\ell(w;z,z')+\lambda \Omega(w)$ is at least $\mu$-strongly convex.
\end{assumption}
\begin{assumption}[Smoothness] 
	\label{ass:smooth} For any $ z,z' \in \mathcal{X}$ the loss function $\ell(\cdot;z,z')$ is L-smooth i.e. there exist a constant $0\leq L\leq \infty$ such that 
	\begin{eqnarray}
	 \|\nabla \ell(w;z,z') - \nabla \ell(w';z,z')\| \leq L \| w- w \|
	\end{eqnarray}
	which is equivalent to :\\
	\begin{equation}
	   \ell(w;z,z') \leq \ell(w';z,z') + \nabla \ell(w';z,z')^T(w-w') + \frac{L}{2}\| w- w' \|^2
	\end{equation}
    Thus we have that $L(w;z,z') :=\ell(w;z,z')+\lambda \Omega(w)$ is at least $L$-smooth function.
\end{assumption}

\section{Methodology}
\label{Methodology}
We first introduce our algorithm for solving the the problem in \ref{eq:ERM} in algorithm \ref{alg:DSGD}. Then, the analysis begins with the double stochastic gradient descent outlined in section 3.1 to solve the subproblems using non-uniform sampling and adatptive sample size. In sections 3.2 and 3.3, we analyze the algorithm's convergence and uniform stability, and in section 3.4, we determine the least number of iterations required to solve any subproblem.

\begin{algorithm}[h]
		\renewcommand{\algorithmicrequire}{\textbf{Input:}}
		\renewcommand{\algorithmicensure}{\textbf{Output:}}
		\caption{Adaptive pairwise learning with DSGD}
		\begin{algorithmic}[1]
			\REQUIRE  Training set $\mathcal{S}$, growth rate $\beta$, initial sample size  $m_0$, the probability vector $P$, step size sequence $\{\gamma_t\}_{t=1}^T$.
			\STATE  Set $m = m_0$, $n = |\mathcal{S}|$ and  initialize   $\bar{w} = w^0$ such that  $R_{m}(w^0) - R_{m}(w^*_{m}) \leq \mathcal{O } \left ( \frac{1}{m^{\alpha}} \right )$.
			\FOR {s = 1 : $\log_\beta{\frac{|\mathcal{S}|}{m_0}}+1 $}
			
			 \STATE $w^0 \gets \bar{w}^{s-1}$
			 \FOR {t=1:m}  
                 \STATE Pick $(x_i,y_i)\sim \mathcal{D}_+$
                 \STATE Pick $(x'_j,y'_j)\sim \mathcal{D}_-$
                \STATE $g_t =  (P(i,j)n(n-1))^{-1}\nabla \ell(w,(x_i,y_i),(x'_j,y'_j)) $
                \STATE $w^{t} = prox_{\gamma_t\lambda,\Omega}(w^{t-1} - \gamma_t g_t)$
             \ENDFOR
			\STATE Set $\bar{w}^s = {w^T}$.
			\STATE $m \gets \min(\beta m,n)$
			\ENDFOR
			\ENSURE $\bar{w}$.
		\end{algorithmic}
		\label{alg:DSGD}
	\end{algorithm}

\subsection{Doubly Stochastic gradient descent (DSGD) with adaptive sample size}

Since the pair data are i.i.d., we expect to have a generalization bound on the difference between the empirical and the true risks, namely; in literature \cite{boucheron2005theory} the bound is found to be function of the sample size as illustrated in (\ref{bound}). 
\begin{equation}
  \label{bound}
\mathbb{E} \left[\sup_{w\in \mathbb{R}^d} |R_n(w) - R(w)|   \right] \leq V_n \approx  \mathcal{O}\left(\frac{1}{n^\alpha}\right)
\end{equation}
In which $\alpha \in [0 \; 0.5]$ corresponds to the algorithm used to solve the ERM and the expectation is set in relation to the training sample
 A small subset of training samples is used to solve the corresponding ERM within its statistical accuracy (based on \ref{bound}). The adaptive sample size training then expands the training samples to include new samples on top of the previously solved ERM and solves the new ERM within its statistical accuracy (defined by the number total samples), where the optimization begins with an initial solution equal to the previous stage solution. The process will continue until all of the samples have been visited at least once. 
\par 
In Theorem \ref{th1}, an upper bound on the empire loss difference of a random stage with $S_n$ dataset is derived .
The theorem, in particular, constrains the absolute difference between the optimal empire risk and the initial empire risk found using a subset $S_m \subset S_n$ in the prior stage. That is, if we define suboptimality of random stage empire loss with $S$ dataset with $m:=|S|$, as the difference $R_n(w_m) - R_n(w^*)$, then the upper bound relies on the optimization accuracy of the ERM solution ($\delta_m$) of the previous stage on $S_m$ and other factors. 

	\begin{theorem}\label{th1}
		Assume that the training set $\mathcal{S}$ is i.i.d.  and we have a subset $\mathcal{S}'$ of $\mathcal{S}$  of the size of $m$, where $\mathcal{S}'$ is drawn from $\mathcal{S}$ with equal probability.
Define $w_m$ as an $\delta_m$ approximate  solution of the risk $R_{m}$ in expectation, \emph{i.e.}, $\mathbb{E} [R_{m}(w_m) - R_{m}(w_m^*)
] \leq \delta_m$, we have that
\begin{eqnarray}\label{th1_1}
 \mathbb{E} \left  [ R_{n}(w_m) - R_{n}(w_{n}^*) \right ]
\leq \delta_m + 2\frac{n-m}{n}   V_{m}
\end{eqnarray}
Proof in appendix.
	\end{theorem}

\begin{remark}
    The ultimate result of Theorem \ref{th1} leads to two important conclusions: 
\begin{itemize}
     \item It is possible to obtain an approximate solution to the ERM by solving the ERM problem on a subset $S'\subseteq S$.
     \item However, even if the most accurate ERM solution on $S'$ exists, i.e. if $\delta_m =0$, the succeeding ERM problem on $S \supset S'$ will always have an optimal solution that is dependent on the $V_m$. Theorem \ref{th1} states that this is true regardless of whether or not $\delta_m =0$. 
 \end{itemize}
\end{remark}
Thus, in light of the above-mentioned findings, we suggest an adaptive technique that includes expanding the training set size.
In spite of the fact that the notion of dynamically growing sample size has been studied in the literature (see e.g. \cite{gu2019scalable,mokhtari2017first}), none of the research currently accessible have dived into the implications of adaptive sample size in non-smooth pairwise learning (i.e. AUC with $l_1$-norm).

If the empire loss in (\ref{eq:ERM}) has a differentiable loss function, then as is the case with univariate loss functions, double stochastic gradient descent seeks to employ an unbiased stochastic gradient.using pair of data. In other words given a sequences of i.i.d. training data $\mathcal{S}:=\{ (x_i,y_i)\}^{n}$ in which $(x_i,y_i) \sim \mathcal{D}$, we define stochastic gradient $g$ of the ERM in \ref{eq:ERM} as:
    \begin{eqnarray}
  g := \nabla \ell(w,(x_i,y_i),(x'_i,y'_i)) 
    \end{eqnarray}
    where $i \sim U(0,n(n-1))$. Assume that variance of $g$ is defined as follow:
    \begin{eqnarray}
    Var[g] = \mathbb{E}_i[\|g-\mathbb{E}_i[g] \|^2]
    \end{eqnarray}
    Before we prove the stochastic gradient used in algorithm \ref{alg:DSGD} have reduced variance compared to uniform distribution we redefine the empirical risk to be unbiased of the true expected risk inspired by \cite{gopal2016adaptive}as illustrated in \ref{eq:ERM2}.
    \begin{equation}
    \label{eq:ERM2}
    R_n(w) := \sum_{k\in\mathcal{C}}\sum_{i\neq j,i,j\in \mathcal{C}_k} P(i,j)\ell(w,z_i,z_j) + \lambda \Omega(w) 
    \end{equation}
    where $P(i,j) = \frac{\mathcal{P}_k}{| \mathcal{C}_k|}$ and $\mathcal{P}_k = \frac{| \mathcal{C}_k|}{n(n-1)}$. Moreover $\mathcal{C}_k$ is a set of pairs defined on $n(n-1)$ such that $\mathcal{C}_k \cap \mathcal{C}_l = \Phi \quad \forall k\neq l$. This concept was developed for the first time by Gopal et al. \cite{gopal2016adaptive} in an effort to reduce variance in pointwise stochastic gradient using class labels. In our pairwise framework, there are four natural sets: two sets with identical classes of the size $n^+(n^+-1)$ and $n^-(n^--1)$ and two symmetric sets with distinct classes of the size $n^+n^-$, where $n^+$ and $n^-$ are the number of positive and negative examples in $\mathcal{S}$. 
    \par 
    We define new distribution over our samples such that $P((x,y=1),(x',y'=0)) = \frac{1}{2n^+n^-}$ and zero otherwise $\forall x,x'\in \mathcal{S}$, by forcing $\mathcal{P}_k=1$ for k=3 and k=4 (sets with opposite labels) and zero for k=1 and k=2. The new gradient is now defined as 
    \begin{eqnarray}
     g_t = (P(i,j)n(n-1))^{-1}\nabla \ell(w,(x_i,y_i),(x'_i,y'_i)) 
    \end{eqnarray}
    In proposition \ref{th:unbiased} we prove that the variance using the new static distribution is less than the uniform distribution gradient.
    
    \begin{proposition}
    \label{th:unbiased}
    Using only opposite pairs to generate the stochastic gradient would results in reduced variance compared to uniform distribution over $[n(n-1)]$ i.e.:
    \begin{eqnarray}
    Var[g_t] < \mathbb{E}_i[\|g-\mathbb{E}_i[g] \|^2]
    \end{eqnarray}
\end{proposition}
Proof in appendix.

\subsection{Uniform Stability}
Solving the optimization problem in adaptive sample size requires crucial definition of the iteration number in every subproblem (or inner loop). We investigate the minimal number of iteration needed in every innep loop using uniform stability and convergence rate of the DSGD, after we define the uniform stability and its relation to generalization error in definitions \ref{def:stability} and \ref{def:gen} (\cite{bousquet2002stability,shen2020stability,agarwal2009generalization}).

\begin{definition}[Algorithmic Stability]
An iterative algorithm $\mathcal{A}:\mathcal{Z}\times \mathcal{Z} \rightarrow \mathbb{R}^d$ is called $\epsilon$-stable w.r.t. the loss function $\ell$ if: 
\begin{equation}
    \label{eq:stability}
    \sup_{i\in[n]}\mathbb{E}| \ell(\mathcal{A}(S),\cdot,\cdot) -\ell(\mathcal{A}(S^i),\cdot,\cdot) | \leq \epsilon \quad \forall S\in \mathcal{Z}^n \quad
\end{equation}
 where the expectation is w.r.t. the algorithm randomness, $[n]:=\{1,\dots,n\}$, and $S^i$ defined in (\ref{eq:modset}). 
\label{def:stability}
\end{definition}
\begin{definition}[Generalization Bound]
Given $\epsilon$-stable algorithm $\mathcal{A}$, the generalization error of the same algorithm given dataset $S$ is bounded above as:
\begin{eqnarray}
  \mathbb{E} [R_n(\mathcal{A}(S))-R(\mathcal{A}(S))] \leq 2\epsilon
  \end{eqnarray}

  which is equivalent to
  \begin{eqnarray}
\mathbb{E} \left[\sup_{w\in \mathbb{R}^d} |R_n(w) - R(w)|   \right] \leq 2\epsilon
\end{eqnarray}
\label{def:gen}
where the expectations is w.r.t. the randomness of algorithm and choice of S.
\end{definition}

In proposition \ref{pro:stability} we introduce the stability bound of our algorithm \ref{alg:DSGD} as a function of the step size and the Lipschits continuity constant.
\begin{proposition}[Stability]
Let w be the output of algorithm 1 on training datasets $\mathcal{S}^+$ and $\mathcal{S}^-$. Let assumption 2 hold with $\mu \geq 0$ and $\gamma_t \leq \frac{1}{L}$, then the algorithm \ref{alg:DSGD} is  $(2G^2 (n(n-1))^{-1} \sqrt{\sum_{t=0}^{T-1}{\gamma_t^2} })$-uniformly stable and further the generalization bound of $w$ is bounded above as:
\begin{eqnarray}
\label{eq:prop1}
\mathbb{E} \left[|R_n(w_T) - R(w_T)|   \right] \leq
4G^2 (n(n-1))^{-1}\sqrt{\sum_{t=0}^{T-1}{\gamma_t^2} }
\end{eqnarray}
\label{pro:stability}
Proof in appendix.
\end{proposition}

The bound of the stability in proposition 1 depends on the sequence of the step size $\{\gamma_t\}$. In corollary \ref{lemma:stepsize} we assumes the standard fixed step size, to have a bound that is dependent on the iteration number $T$.
\begin{corollary}
If the step size is chosen to be constant i.e. $\gamma_t = \gamma $ $\forall t=1,\dots,T$, then the bound in proposition \ref{pro:stability} is :
\begin{eqnarray}
\mathbb{E} \left[|R_n(w_T) - R(w_T)|   \right] \leq
4G^2 \gamma (n(n-1))^{-1} \sqrt{T}
\end{eqnarray}
\label{lemma:stepsize}
\end{corollary}
In the next section we derive the convergence rate for algorithm \ref{alg:DSGD} and together with the stability bound we can arrive at the optimal iteration number needed in the inner loop to have a solution that is statistically accurate.

\subsection{Convergence Analysis}
\begin{theorem}
\label{th:convergence}
The DSGD algorithm under assumptions \ref{ass:convex},\ref{ass:smooth} and bounded variance of gradient e.g.:
\begin{eqnarray}
\mathbb{V}[g_t]:=\mathbb{E}\|\nabla \ell_{i,j}(w,x_i,x_j) - \nabla \ell(w,x_i,x_j)\|^2 \leq \sigma^2
\end{eqnarray}
have sublinear convergence rate $O(1/T)$ given the proximal operator of $ \Omega(w)$ is cheaply computed, i.e. given an initial solution such that $w_s^0 = w_{s-1}$ we have for any stage $s$ in DSGD algorithm \ref{alg:DSGD}:
\begin{align}\mathbb{E}[R(w_s^{T})- R(w^{*})] 
\leq   \frac{1}{\gamma_t  \mu T} (R(w_{s-1})- R(w^{*}))+  \gamma_t\sigma^2
      \label{eq:bound6}
\end{align}
	where $\gamma >0$ depends on the setting of the loss function.
\end{theorem}

\begin{remark}
The convergence and stability bound of DSGD indicates the following:
\begin{enumerate}
    \item Variance of the gradient $\sigma^2$ could slow the convergence.
    \item The generalization bound indicates the iteration number expand the bound $O(\sqrt{T})$.
\end{enumerate}
\end{remark}
\subsection{Computational Complexity}
 Based on proposition \ref{pro:stability} and theorem \ref{th:convergence}, we chose the number of iteration in every inner loop by minimizng both bounds in $T$ as follow:
 \begin{eqnarray}
 T^* = \argmin_T \frac{1}{\gamma_t  \mu T} + 4G^2 \gamma (n(n-1))^{-1} \sqrt{T}
 \end{eqnarray}
 The solution can be found easily as $T^*=O(n(n-1))^{2/3}) = O(n^{4/3})$.


\par

Theorem \ref{th:comp} states the computational complexity of our algorithm \ref{alg:DSGD} is $O(n)$ where n is the number of examples, and the complexity is expressed in terms of gradient components to reach $\epsilon$-optimal solution of the ERM. 
\begin{theorem}
\label{th:comp}
    Finding $V_n$- statistical accurate solution of the ERM problem in \ref{eq:ERM} using algorithm \ref{alg:DSGD}, under the assumptions \ref{ass:convex},\ref{ass:smooth}, and theorem \ref{th:convergence}, will have a computational complexity given by:
    \begin{eqnarray}
    \sum_{s=1:\log_{\beta}\frac{n}{m_0}} O(m_s) = O\left(\log_{\beta}\frac{n}{m_0}\right) m_s \approx O(n)
    \end{eqnarray}
    where $\beta$ is the expansion factor as defined by algorithm \ref{alg:DSGD}, $m_0$ intial sample size, and n is the total sample size.
\end{theorem}

\begin{table}[!t]
  	\small
 \center
\caption{Datasets names, features and number of examples used in the experiments}
\begin{tabular}{ccccc}

\hline
\multirow{2}{*}{} &
\multirow{2}{*}{Dataset} & \multirow{2}{*}{Examples} & \multirow{2}{*}{Features} & \multirow{2}{*}{Reference} \\
                &         &                           &                           &                                        \\ \hline
1 & a9a                      & 32,561                    & 123                       & \cite{Dua:2019}       \\2&
rcv1                     & 20,242                    & 47,236                    & \cite{lewis2004rcv1}  \\3&
fourclass                & 862                       & 2                         & \cite{ho1996building} \\4&
diabetes                 & 768                       & 8                         & \cite{Dua:2019}       \\5&
covtype                  & 581,012                   & 54                        & \cite{Dua:2019}       \\6&
ijcnn1                   & 49,990                    & 22                        & \cite{CC01a}
\\
\hline
\label{datasets}

\end{tabular}
\end{table}

\section{Experiments}
\label{sec:exp}
The experiments in this part are focused on pairwise learning in AUC maximization, which is the approach that attempts to maximize the area under the receiver operating characteristic curve (ROC) through a variety of ways but not including the pointwise learning.

\subsection{Problem Formulation}
 AUC metric maximization is defined as in equation (\ref{eq:AUC}) as first introduced in \cite{hanley1982meaning} that measures the probability of obtaining higher weight on the positive label. 
Given a space $\mathcal{X}\times \mathcal{Y}$ with unknown distribution $\mathcal{P}$ and $\mathcal{Y} = \{+1,-1\}$. A function that takes a sample,e.g. $x\in \mathcal{X}$ drawn independently according to $\mathcal{P}$ and predicts the classes $f:\mathcal{X} \rightarrow \mathcal{Y} $ have AUC score given by:
\begin{align}
    AUC(f) :=& Pr(f(x)>f(x')|y=1,y'=-1)  = \mathbb{E}[\mathbb{I}_{f(x)>f(x')}|y=1,y'=-1]
    \label{eq:AUC}
\end{align}
where the expectation is w.r.t. the samples. However
 given the fact that the formulation in (\ref{eq:AUC})  is neither convex nor differentiable, researchers have approximated the AUC by using surrogate convex functions, such as the square function and hinge loss. In addition to the fact that data are often limited and the distribution is unidentifiable, it is impossible to calculate the expectation directly; as a result, we present the empirical AUC score in (\ref{eq:EAUC}) with linear model and regularized squared loss function.
 
 \begin{table}[!b]
\centering
\caption{AUC maximization results (average  $\pm$ standard error) using different batch and online algorithms on different datasets}
\label{AUC_results}
\begin{tabular}{@{}llllllll@{}}
\toprule
\centering
\textbf{DS} & \textbf{adaDSGD}                                       & \textbf{\begin{tabular}[c]{@{}l@{}}SPAM\\ NET\end{tabular}} & \textbf{SOLAM}                                         & $\bm{OAM_{SEQ}}$                                       & $\bm{OAM_{GRA}}$                                       & \textbf{B-LS-SVM}                                      & \textbf{OPAUC}                                         \\ \midrule
1           & \begin{tabular}[c]{@{}l@{}}.8998\\ ±.0034\end{tabular} & \begin{tabular}[c]{@{}l@{}}.8995\\ ±.0042\end{tabular}      & \begin{tabular}[c]{@{}l@{}}.9001\\ ±.0042\end{tabular} & \begin{tabular}[c]{@{}l@{}}.8420\\ ±.0174\end{tabular} & \begin{tabular}[c]{@{}l@{}}.8571\\ ±.0173\end{tabular} & \begin{tabular}[c]{@{}l@{}}.8982\\ ±.0028\end{tabular} & \begin{tabular}[c]{@{}l@{}}.9002\\ ±.0047\end{tabular} \\ \midrule
2           & \begin{tabular}[c]{@{}l@{}}.9938\\ ±.0009\end{tabular} & \begin{tabular}[c]{@{}l@{}}.9878\\ ±.0015\end{tabular}      & -                                                      & -                                                      & -                                                      & -                                                      & -                                                      \\ \midrule
3           & \begin{tabular}[c]{@{}l@{}}.4343\\ ±.1783\end{tabular} & \begin{tabular}[c]{@{}l@{}}.8231\\ ±.0445\end{tabular}      & \begin{tabular}[c]{@{}l@{}}8226\\ ±.0240\end{tabular}  & \begin{tabular}[c]{@{}l@{}}.8306\\ ±.0247\end{tabular} & \begin{tabular}[c]{@{}l@{}}.8295\\ ±.0251\end{tabular} & \begin{tabular}[c]{@{}l@{}}.8309\\ ±.0309\end{tabular} & \begin{tabular}[c]{@{}l@{}}.8310\\ ±.0251\end{tabular} \\ \midrule
4           & \begin{tabular}[c]{@{}l@{}}.8284\\ ±.0311\end{tabular} & \begin{tabular}[c]{@{}l@{}}.8203\\ ±.0324\end{tabular}      & \begin{tabular}[c]{@{}l@{}}.8253\\ ±.0314\end{tabular} & \begin{tabular}[c]{@{}l@{}}.8264\\ ±.0367\end{tabular} & \begin{tabular}[c]{@{}l@{}}.8262\\ ±.0338\end{tabular} & \begin{tabular}[c]{@{}l@{}}.8325\\ ±.0329\end{tabular} & \begin{tabular}[c]{@{}l@{}}.8309\\ ±.0350\end{tabular} \\ \midrule
5           & \begin{tabular}[c]{@{}l@{}}.8238\\ ±.0019\end{tabular} & \begin{tabular}[c]{@{}l@{}}.6828\\ ±.0016\end{tabular}      & \begin{tabular}[c]{@{}l@{}}.9744\\ ±.0004\end{tabular} & \begin{tabular}[c]{@{}l@{}}.7361\\ ±.0317\end{tabular} & \begin{tabular}[c]{@{}l@{}}.7403\\ ±.0289\end{tabular} & \begin{tabular}[c]{@{}l@{}}.8222\\ ±.0014\end{tabular} & \begin{tabular}[c]{@{}l@{}}.8244\\ ±.0014\end{tabular} \\ \midrule
6           & \begin{tabular}[c]{@{}l@{}}.9316\\ ±.0033\end{tabular} & \begin{tabular}[c]{@{}l@{}}.8715\\ ±.0101\end{tabular}      & \begin{tabular}[c]{@{}l@{}}.9215\\ ±.0045\end{tabular} & \begin{tabular}[c]{@{}l@{}}.7113\\ ±.0590\end{tabular} & \begin{tabular}[c]{@{}l@{}}.7711\\ ±.0217\end{tabular} & \begin{tabular}[c]{@{}l@{}}.8210\\ ±.0033\end{tabular} & \begin{tabular}[c]{@{}l@{}}.8192\\ ±.0032\end{tabular} \\ \midrule
Reg         & $l_1+l_2$                                              & $l_1+l_2$                                                   & $l_2$                                                  & $l_2$                                                  & $l_2$                                                  & $l_2$                                                  & $l_2$                                                  \\
Ref         & Ours                                                   & \cite{pmlr-v80-natole18a}                  & \cite{ying2016stochastic}             & \cite{zhao2011online}                 & \cite{zhao2011online}                 & \cite{joachims2006training}           & \cite{gao2013one}                     \\ \bottomrule
\end{tabular}
\end{table}
\begin{align}
    AUC(w) = \frac{1}{2n^+ n^-} \sum_{i\in[n^+],j\in[n^-]} (1 - [w^T(x_i^+  - x_j^-)])^2 + \lambda \Omega(w)
    \label{eq:EAUC}
\end{align}
where $x_i^+$, $x_i^-$ denote the positive and negative examples respectively, $w\in \mathbb{R}^d$ is d-dimensional linear model weight and $\Omega(w)$ is non-smooth but convex regularization. The elastic net  is considered i.e. $\Omega(w) = \lambda \|w\|^2 + \lambda_1 \|w\|_1$, to have fair compassion with AUC maximization algorithms in literature. However, $l_1$ (lasso) or mixture of both $l_1$ and $l_2$ (group lasso) can be applied. The proximal step (part 8) in algorithm \ref{alg:DSGD} for elastic net can be easily computed given $z$ the stochastic gradient step 7, as follow:
\begin{eqnarray}
&prox_{\gamma ,\Omega}\left(\frac{1}{2\gamma}\|w-z\|^2 + \lambda \|w\|^2 + \lambda_1 \|w\|_1 \right)  =prox_{\gamma ,\Omega}\left( \frac{1}{2\gamma} \left\| w - \frac{z}{\gamma \lambda +1} \right\|^2 + \frac{\lambda_1}{\gamma \lambda +1}\|w\|_1 \right)
\end{eqnarray}
where $\lambda$, $\lambda_1$ are the regularization weights, and $\gamma$ is the step size.

\begin{figure}[!h]
\centering
\label{fig}
\includegraphics[width = \textwidth]{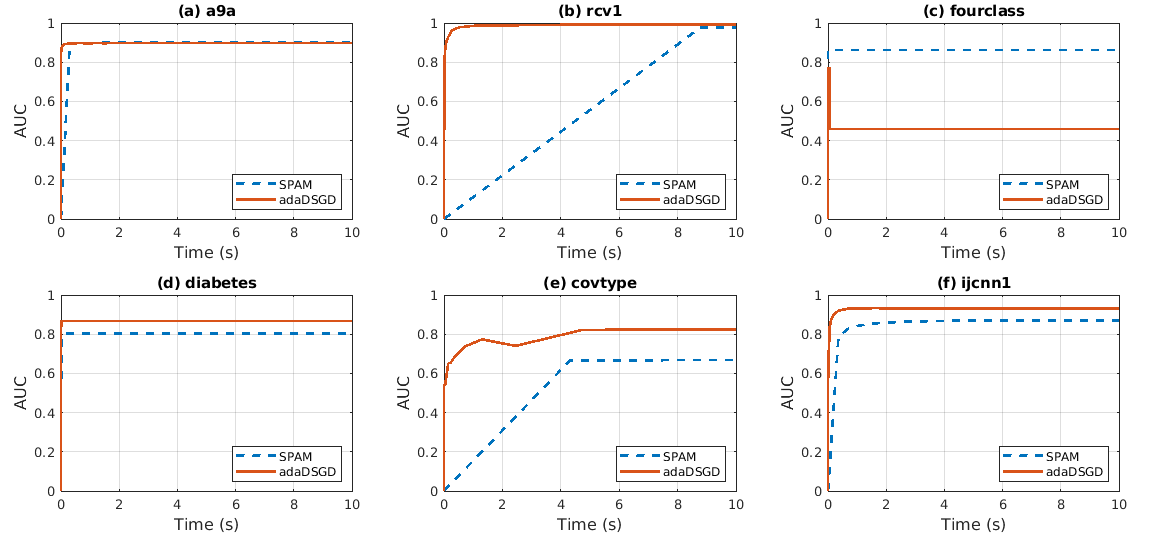}
 \caption{AUC maximization using adaptive double stochastic gradient descent (adaDSGD) algorithm and stochastic proximal AUC maximization (SPAM) algorithm) on six datasets in terms of computational time.}
\end{figure}

\subsection{Datasets}
The datasets names used in this experiments are listed in table \ref{datasets} along with their features, sample size, and source reference. Since AUC maximization approach is based on binary classification, where nonbinary datasets are converted to binary by randomly dividing the labels into two groups. Every dataset is divided into two parts: 80 percent training and 20 percent testing; in addition, the experiments on each dataset are repeated 25 times with different random seeds, and the standard error and mean are calculated for each experiment. The experiments are conducted in MATLAB on intel 2.4 GHz machine with 4 GBs RAM.
\par
Experiment findings are separated into two parts: first, we compare the complexity of the algorithm as measured by its running time to SPAM-NET since, as shown in its experiment section, it is the superior algorithm \cite{pmlr-v80-natole18a}. In the second part of this analysis, we evaluate the outcomes of the AUC in terms of the arithmetic mean and the standard error of each method displayed in table \ref{AUC_results}. Both aspects of the study demonstrate that our algorithm is better in terms of the AUC findings and the length of time it takes to execute (except for dataset number 3 fourclass).


\section{Conclusions}
\label{sec:con}
In this study, a new approach for pairwise objective paradigms was investigated. A double stochastic gradient descent (DSGD) in stagewise phase is presented as a workable solution of sample size, which increases quadratically with the number of samples. In particular , the training set is divided into finite number of smaller sets, and every outerloop the algorithm expand the training set to include new set. Moreover, we used convergence bound and uniform stability of DSGD to calculate the lowest number of iterations required for each stage.
Concurrently, we presented a new distribution over the different sample space in order to lower the gradient variance induced in DSGD, by sampling opposite instances at each iteration.
AUC maximization experiments conducted on a wide range of datasets confirm the theoretical predictions. It is possible to do further investigation in order to develop a more sophisticated, non-static importance sampling strategy.

\bibliographystyle{unsrt} 
\bibliography{Final_version}

\end{document}